\documentclass{article}
\usepackage{spconf,amsmath,graphicx}

\usepackage{algorithm}
\usepackage{algorithmic}
\usepackage{amsfonts, bm}

\usepackage[colorlinks = true,
            linkcolor = blue,
            urlcolor  = blue,
            citecolor = blue,
            anchorcolor = blue]{hyperref}
            
\usepackage{tabularx, booktabs}
\usepackage{xcolor}

\def\twosplit#1{%
\clearpage
\def\row##1##2{##1}%
#1%
\def\row##1##2{##2}%
#1%
}

\title{Deep Spatio-Temporal Wind Power Forecasting}
\name{Jiangyuan Li and Mohammadreza Armandpour}
\address{Department of Statistics, Texas A\&M University, College Station, TX}
\begin{document}
%
\maketitle
\begin{abstract}
Wind power forecasting has drawn increasing attention among researchers as the consumption of renewable energy grows. In this paper, we develop a deep learning approach based on encoder-decoder structure. Our model forecasts wind power generated by a wind turbine using its spatial location relative to other turbines and historical wind speed data. In this way, we effectively integrate spatial dependency and temporal trends to make turbine-specific predictions. 
The advantages of our method over existing work can be summarized as 1) it directly predicts wind power based on historical wind speed, without the need for prediction of wind speed first, and then using a transformation; 2) it can effectively capture long-term dependency 3) our model is more scalable and efficient compared with other deep learning based methods. The implementation and trained models are available on \url{https://github.com/jiangyuan2li/Deep-Spatio-Temporal}. We demonstrate the efficacy of our model on the benchmark real-world datasets. 
\end{abstract}

\section{Introduction}
Wind energy has become an essential source of energy resources worldwide due to being pollution-free, and its wide availability \cite{wiser2015wind}. However, its strong volatility could cause substantial power fluctuation and affect the overall operation of the regional power grid. Insufficiently accurate wind forecasts may bring hidden dangers to the safe and stable operation of the entire power system. Therefore, an effective wind power forecasting method is necessary to find the most economical solution for the operation of the power grid. This can help the power dispatching department to organize the generation plan optimally and consequently improves the reliability and security of the power grid \cite{chen2013wind, azad2014long}.

Wind power forecasting is generally viewed as a complex task due to the chaotic and stochastic features of wind speed time series. Existing methods on wind power forecasting fall into four main categories.
1) Persistence methods assume the wind data remain unchanged in a short time window.
2) Physical methods formulate the problem of wind power based on numerical weather prediction (NWP) usually use weather prediction data such as temperature, pressure, surface roughness, and obstacles. NWP build models by complete hydrodynamic and thermodynamic equation sets, which usually have huge computational burdens but with limited temporal and spatial resolution~\cite{soman2010review}.
3) Statistical methods are based on probabilistic modeling on historical data, such as ARIMA-based approaches for temporal features~\cite{erdem2011arma} and Kriging interpolation method for spatial correlations.
4) Deep learning based methods learn the intricate mapping between the inputs and outputs from massive historical data.

The recently developed deep learning approach shows a significant improvement compare to the classical baseline. However, most of them still suffer from several drawbacks: 1) They lack a proper design for temporal features and spatial correlations. 2) Many of them are still relying on power curve transformation to forecast wind power output. Although this approach could simplify the wind power forecasting problem to wind speed time series analysis, the power curve-fitting leads to considerable errors, which also leaves turbine identity aside. 3) Capturing long-term dependency in current methods requires having a large neural network with many parameters, which is not data-efficient and may not be a necessity. Furthermore, enlarging the network size may lead to severe overfitting issues and cause difficulties in dealing with strong seasonality due to the nature of the wind. 

To overcome these problems, we develop a deep wind power forecasting model. Our model adopts an encoder-decoder architecture with GRU \cite{chung2014empirical} as the recurrent unit, which can capture the temporal feature with long-term dependency. With the extra multi-layer perceptron (MLP) attached in the decoder, the model could directly produce forecasts for either wind speed or wind power. Moreover, we utilize spatial information of turbines and the correlation among neighbor turbines to provide a more robust and accurate forecast. We also learn an embedding vector to produce turbine-specific forecasts, which account for each turbine's quality and unique environmental condition. 

\section{Problem Formulation}
The objective of wind power forecasting is to capture the relations between the historical wind speed data of surrounding wind turbines and the future wind power output of the target wind turbine. Suppose there exists a set of $N$ wind turbines in a wind farm, each wind turbine is determined by its locations $\nu^i\in V$. From the historical data, the wind speed at time $t$ is denoted by $X_t = (x_t^1, x_t^2, \ldots, x_t^N)$ and the recorded wind power output is denoted by $Y_t = (y_t^1, y_t^2, \ldots, y_t^N)$. At time $t$, for our target turbine $i$, the forecast $\hat{y}_{t+\tau}^i$ is predicted based on the $m$ previous wind speed data, i.e.,
\begin{equation}
\begin{aligned}
    h_t &= f_e(X_{t-m+1},X_{t-m+2},\cdots, X_t|\Phi_e)\\
    \hat{y}_{t+\tau}^i &= f_d(y_t, h_t|\Phi_d)
\end{aligned}
\end{equation}
where $\tau$ denotes the forecasting horizon, $f_e$ is an implicit function to extract hidden features for forecasting, $f_d$ is for making forecasts, which takes current wind power record $y_t$ and hidden states $h_t$ as inputs. $\Phi_e$ and $\Phi_d$ represents the parameters in our model. Note that ${X_t}$ can be viewed as a multi-dimensional time series. The proper choice of dimension is determined by the spatial correlations. For each turbine $i$, we build the neighbor and denoted as $k(i)$. Many other features besides wind speed can also be incorporated in this multi-dimensional time series and will be discussed later. For model parameters $\Phi$, we separate one set of parameters $\theta_i$ for each turbine's identity to make turbine-specific forecasts. Thus, the forecasting framework takes the following form,
\begin{equation}
\begin{aligned}
     h_t^i &= f_e(X^{k(i)}_{t-m+1},X^{k(i)}_{t-m+2},\cdots, X^{k(i)}_t|\Phi_e,\theta_i)\\
     \hat{y}_{t+\tau}^i &= f_d(y_t, h_t|\Phi_d, \theta_i).
\end{aligned}
\end{equation}
Note that the parameters $\Phi_e$, $\Phi_d$ and $\theta_i$ are trained together, and $\theta_i$ as turbine identity only serves for our forecasting purpose.
\section{Proposed Methodology}
In this section, the extraction of deep spatio-temporal features of the proposed model is explained. First, the temporal feature is extracted by gated recurrent unit (GRU), and worked in an encoder-decoder manner which transforms historical wind speed to wind power forecasts. Second, the temporal features are enriched by spatial correlations using graph construction. Last, turbine identity embedding and relevant time features are included in the model to produce turbine-specific forecasts and enhance the model performance.

\subsection{Temporal Features}

Assume the wind speed values corresponding to time steps $\tilde{t}\leq t$ are available at time step $t$. We consider the $m$-length time window $[t-m+1,\ldots,t-1, t]$ to capture the temporal wind speed data features. For wind turbine $i$, we aim to extract the corresponding hidden features $h_t^i$, and use the hidden features to make multiple step forecasts $\hat{y}^i_{t+\tau}$ up to the given forecasting horizon. In the proposed framework, an encoder-decoder GRU network is applied to learn deep temporal features $h^i_t$. The encoder and decoder are both GRU networks and take inputs sequentially, which captures the sequential nature of the wind time series.

\begin{figure}
  \centering
  \includegraphics[width = \linewidth]{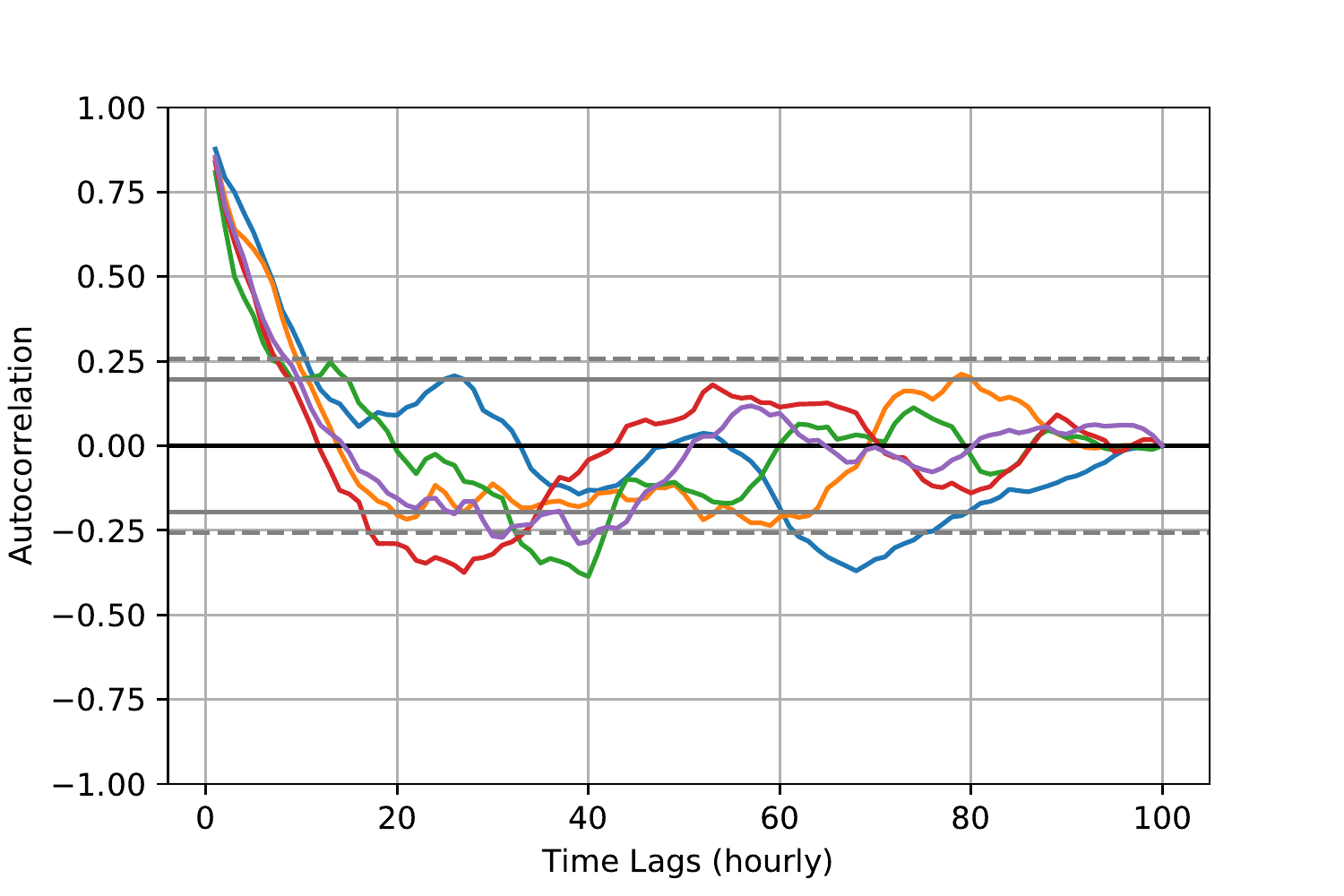}
  \caption{Autocorrelation plots of randomly sampled time series. The dashed and solid horizontal lines indicate to the $99\%$ and $95\%$ confidence intervals for the correlation values around zero.}
  \label{autocor}
\end{figure}

The GRU network is a variation of LSTM network, which has less parameters and can be trained faster. The GRU block has several special multiplicative computational units. The reset gate controls the output flow of the previous hidden states into the subsequent memory block. The update gate controls the balance between previous hidden states and the current candidate hidden states. GRU utilizes hidden states efficiently instead of using an extra cell state to account for long-term dependency. We randomly pick several wind speed time series and illustrated that the wind speed time series doesn't exhibit extremely long dependencies, see Figure \ref{autocor}. Less parameters in GRU not only benefit the training speed, but also reduce the effect of potential overfitting issues. More discussions about LSTM and GRU are shown in experiments.

\subsection{K-NN Graph}
Analysts have noticed that valuable information may be revealed by considering spatial measurements in a local region, as wind characteristics at a site may resemble those at neighboring sites \cite{ding19}. Deep learning for spatio-temporal data has been widely applied in various spatio-temporal data mining tasks such as predictive learning, representation learning and anomaly detection \cite{wang2019deep}. There are various types of spatio-temporal data that differs in the way of data collection and representation in different applications. For example, convolutional neural network (CNN) is often used in traffic prediction problem to process their image-like data for spatial relationships \cite{yao19}. Attention mechanism is also applied in wind power forecasting across wind farms, in which the geographical coordinates of wind farms can not provide clear information for wind power forecasting patterns \cite{fu2019spatiotemporal}.

Our model tries to make wind power forecasts at a turbine level. Turbines in a local region may share similar air density, air pressure and humidity. Including turbines with similar conditions is beneficial to making forecasts. Distance of turbines provides a natural metric to quantify similarities. To incorporate the spatial dependency, we apply the k-nearest neighbors algorithm (k-NN) on the geographical coordinates of turbines. By incorporating neighbors of the target turbine, the input of encoder would be k-dimensional corresponding to a multidimensional time series, while the decoder remains to generate wind power forecasts for the target turbine. Let $k(i)$ be the index set of k-nearest neighbours of turbine $i$, and $x^{k(i)}_t$ be the wind speed of these $k$ turbines at time $t$, ordered by distance. The objective function would be 
\begin{equation}
    \max_\Phi \log p(\bm{y^i}|\{x_1^{k(i)},\cdots,x_T^{k(^i)}\},\Phi).
\end{equation}

Here, we simply denote all trainable parameters as $\Phi$. When the spatial dependency is unclear, self-attention mechanism \cite{fu2019spatiotemporal} is often used, and it provides a weighted one-dimensional time series as input. This approach explores unknown relationships across wind farms, but possibly is very restrictive. In our model, distance is better used to quantify spatial dependency and we enrich the single time series to k-dimensional, which also provides more flexibility to encode wind speed information in encoder state $z$.

\subsection{Turbine Embedding}
In the wind industry, a power curve \cite{lee2015power} is often used to assessing a turbine’s energy production efficiency, which is relation between the power output and wind speed at the same time. Classical approaches \cite{aziz18, aziz19, pourhabib2016} to get forecasting on wind power rely on this power curve to transform the predicted wind speed to the power output, where the prediction on wind speed is obtained from a time series or spatial temporal model. There is a large discrepancy of power curves across turbines. The power curve is estimated for each turbine, therefore the forecasting would be turbine-specific. This approach suggests wind speed is the main feature when predicting wind power. The problem is that this extra curve fitting step might add error for wind power forecasting. That also motivates us to use a Seq2Seq approach.

To accomplish the turbine-specific forecasting while allowing the model to share parameters across turbines, we need to give the model information about which turbine the data comes from via input. One hot-encoded vector is a traditional way to identify turbines. But this representation is large, sparse and inefficient without any semantic information. To overcome this inefficiency, we represent each turbine with a latent vector. There is a wide range of topics to use latent vectors or embeddings to represent identities in the model, such as social network analysis \cite{hamilton2017inductive} and natural language processing \cite{mikolov2013distributed}. The embedding vector is usually from feature engineering, or pre-trained embeddings for other tasks. It can also be learned by adding an embedding layer to the model. We follow this learned embedding approach and denote the embedding vector as $g(i)$.

\begin{figure}
  \centering
  \includegraphics[width=\linewidth]{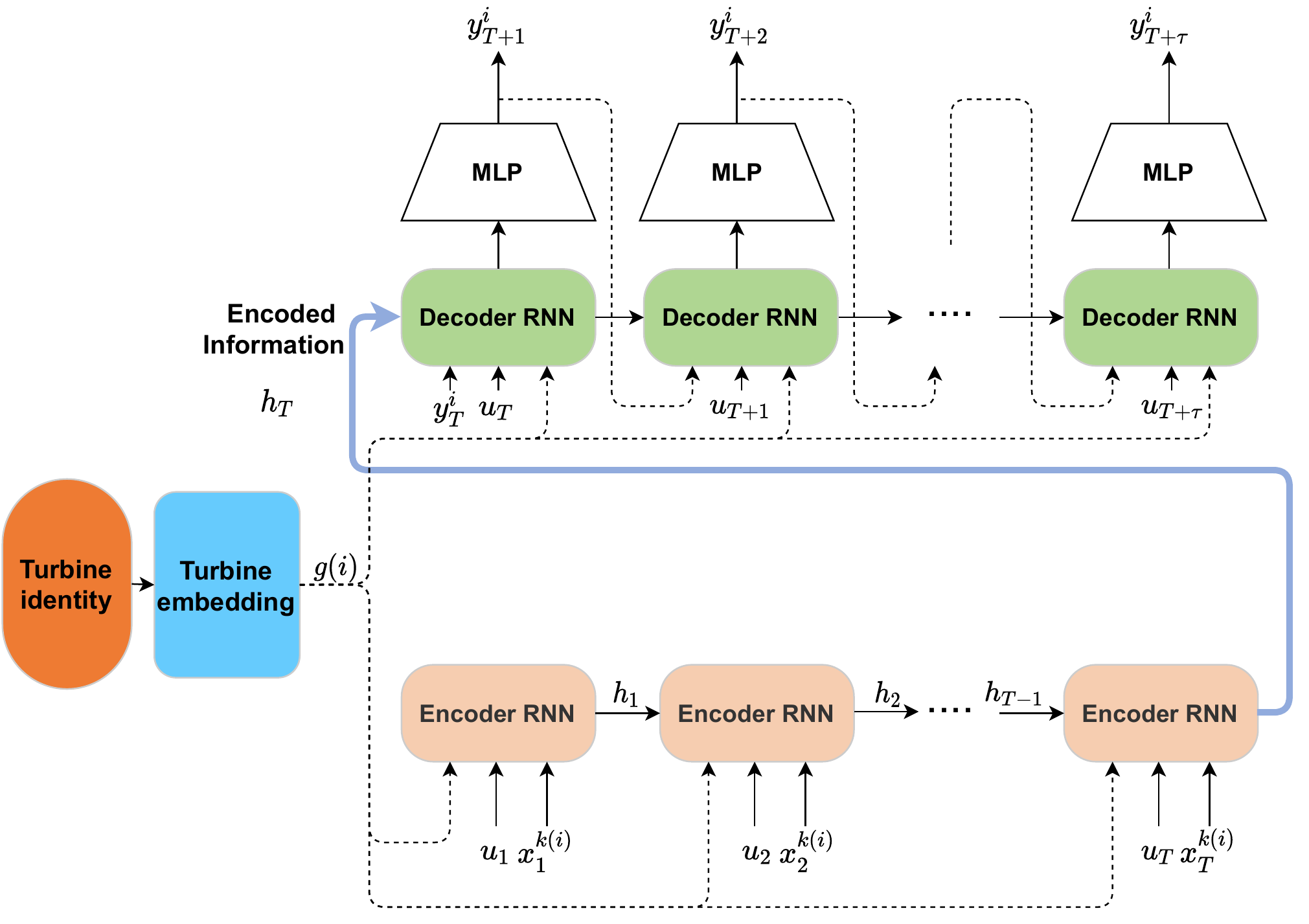}
  \caption{The graphical illustration of the proposed model}
  \label{finalModel}
\end{figure}

The final model is shown in Figure \ref{finalModel}. The embedding layer reads
\begin{equation}
g(i) = E\bm{e}_i,\quad E\in\mathbb{R}^{d_{E}\times d_{i}}.    
\end{equation}
The embedding matrix is denoted as $E$. We denote the dimension of embedded vector as $d_E$, while $d_\nu$ is the number of turbines and $\bm{e}_\nu$ is the one-hot encoded vector of turbine $\nu$. The similarity of turbines is also revealed in the embeddings, which helps capture the spatial dependency. 
\subsection{Feature Enrichment}
Wind speed is known to have seasonal changes. It also shows a daily pattern, which is closely related to sunrise and sunset. In classical approaches, this periodic pattern is often ignored \cite{erdem2011arma, sideratos12}.  Our model can be easily extended to include these time features. Let us denote $u_T$ be the time features, such as hour, month and season at time $T$. We append these time features to the input of each recurrent unit in our model, which further increases the dimension of the input time series. This makes the latent features $h_t$ informative to time, and also helps reduce the influence of change points when dealing with temporal features.

\section{Experiments}
\begin{table}
    \centering
\twosplit{
\begin{tabular}{lcccccccccccc}
\toprule
\row{Time (h) &    1 &    2 &    3 &    4 &    5 &    6 }{&    7 &    8 &    9 &   10 &   11 &   12 }\\
\midrule
\row{PER    &  .128 &  .163 &  .189 &  .212 &  .229 &  .241 }{PER&  .254 &  .268 &  .285 &  .295 &  .299 &  .296} \\
\row{CRS    &  .125 &  .159 &  .185 &  .202 &  .215 &  .223 }{CRS&  .236 &  .241 &  .250 &  .255 &  .260 &  .262} \\
\row{MLP    &  .131 &  .162 &  .184 &  .201 &  .214 &  .223 }{MLP&  .230 &  .236 &  .253 &  .257 &  .259 &  .263} \\
\row{RNN    &  \textbf{.123} &  .155 &  .178 &  .195 &  .209 &  .220 }{RNN&  .230 &  .237 &  .243 &  .248 &  .259 &  .263} \\
\row{LSTM   &  \textbf{.123} &  .157 &  .180 &  .198 &  .255 &  .262}{LSTM &  .263 &  .269 &  .271 &  .270 &  .272 &  .273} \\
\row{PSTN   &  .125 &  .165 &  .177 &  .198 &  .217 &  .233 }{PSTN&  .241 &  .236 &  .245 &  .251 &  .254 &  .253} \\
\row{DL-STF &  .130 &  .161 &  .183 &  .196 &  .208 &  .218 }{DL-STF&  .228 &  .236 &  .243 &  .247 &  .255 &  .256} \\
\row{STAN   &  .132 &  \textbf{.154} &  \textbf{.173} &  .190 &  .203 &  .215 }{STAN&  .223 &  .231 &  .238 &  .244 &  .249 &  .253} \\
\row{Ours   &  .128 &  .155 &  .174 &  \textbf{.189} &  \textbf{.202} &  \textbf{.212} }{Ours&  \textbf{.220} &  \textbf{.227} &  \textbf{.233} &  \textbf{.237} &  \textbf{.242} &  \textbf{.245}} \\
\bottomrule
\end{tabular}
}

    \caption{MAE for wind power forecasting for $h$-hour ahead, $h=1,2,\ldots,12$.}
    \label{res_table}
\end{table}

In this section, we illustrate the performance of our model on two real world datasets. The first one is collected at an onshore wind farm in the United States. One year of turbine-specific hourly wind speed and power values are measured on each of the 200 turbines. This dataset is provided in \cite{ding19}. The other one is from Wind Integration National Dataset (WIND), provided by the National Renewable Energy Laboratory (NREL) \cite{draxl2015overview}. Base on WIND, a $10\times 10$ wind turbine array within a wind farm in Wyoming is selected.

We compare our model with several methods ranging from classical models and the most recent deep learning based models. The persistent model (PER) provides a baseline, and calibrated regime-switching model (CRS) achieved the best performance among several traditional methods \cite{aziz19}. Three base deep learning models are included in comparison, multi-layer perceptron (MLP), recurrent neural net and long-short term memory (LSTM) \cite{hochreiter1997long}. We also compare our model with several other deep learning models for wind forecasting, PSTN \cite{zhu2019learning}, DL-STF \cite{ghaderi2017deep} and STAN \cite{fu2019spatiotemporal}. We use mean absolute error (MAE) and root mean squared error (RMSE) for evaluation purpose.

The result in MAE for wind power forecasting is shown in Table \ref{res_table}. The training period is first 3 months and testing period is the remaining 9 months. Vanilla RNN and LSTM performs the best for the first hour forecasting (better than persistent model), which indicates simple RNN structure is efficient to capture the temporal dependency. Our model obtain the nearly best result for $h=1,2,3$. Starting from $h=4$, our model beats all the methods, which demonstrate the ability of our method to capture the long-term dependency. The number of parameters in our model is around $20k$, while other deep learning methods (PSTN, DL-STF and STAN) need at least $100k$ parameters. This indicates that proper design of deep learning methods is able to capture the spatio-temporal dependency in wind data, and model complexity doesn't need to be unnecessarily large (Table \ref{tab:num}).

\begin{table}
    \centering
\begin{tabular}{ccccccc}
\toprule
{} & PSTN & DL-STF &STAN&Ours\\
\midrule
\#Param & 2.20M  &  17.59M & 225.23M  &  22.40K\\
\bottomrule
\end{tabular}
    \caption{Number of parameters}
    \label{tab:num}
\end{table}

\begin{figure}
  \centering
  \includegraphics[width = 3in]{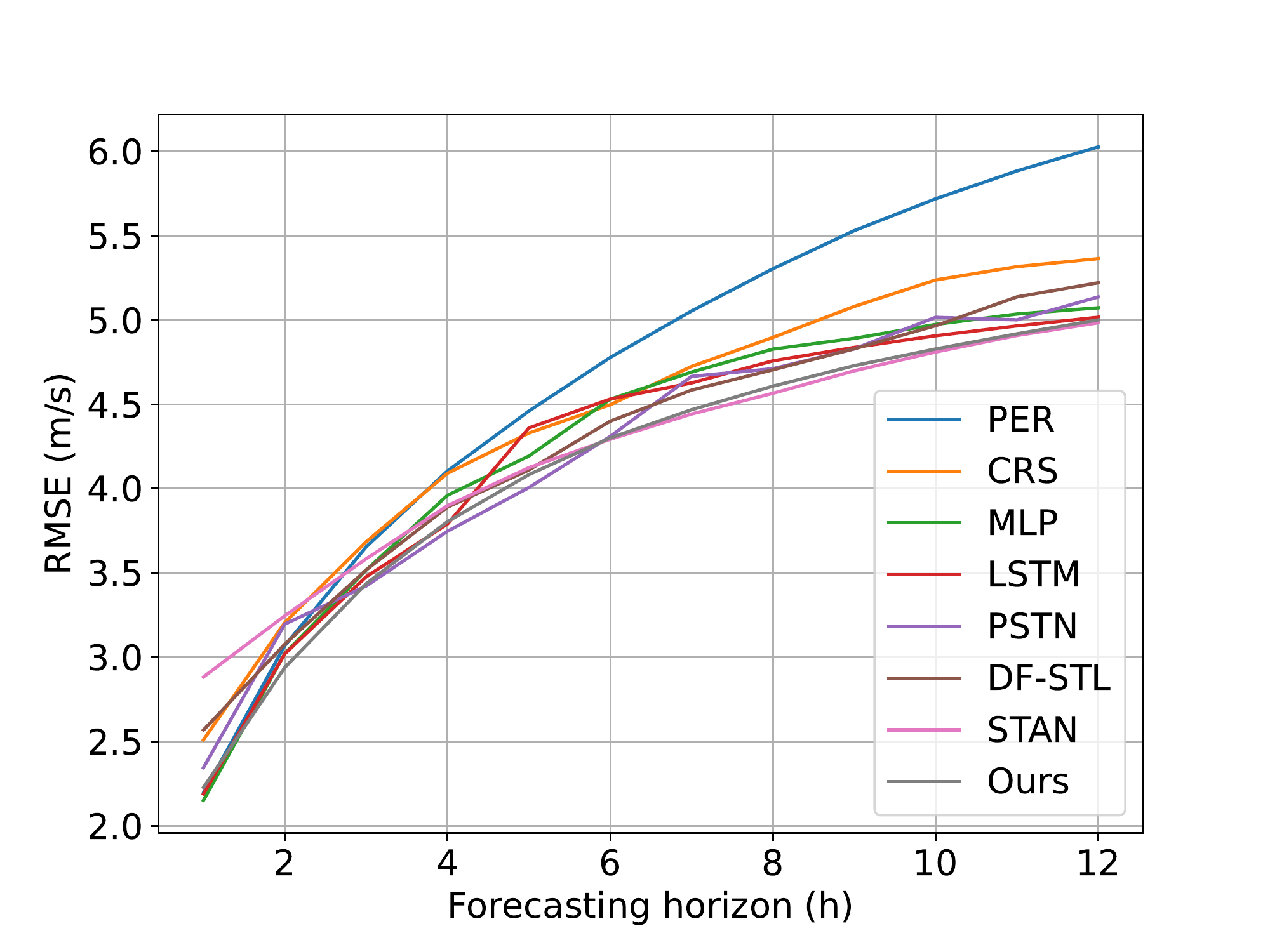}
  \caption{RMSE for wind speed forecasting on NREL dataset.}
  \label{rmse_plot}
\end{figure}

Though wind power data is not available in WIND, we train our model with wind speed to make speed forecasts and compare it with other models. The result in RMSE is shown in Figure \ref{rmse_plot}. The training period is first 8 months and testing period is set as the remaining 4 months to align with \cite{zhu2019learning}. The error curve of our model is consistently lower than all other models, though our model is not designed for speed forecasting. This shows that our model is efficient to extract features of chaotic and stochastic time series of wind data.

\section{Conclusion}
In this work, we proposed a deep spatio-temporal learning approach for wind power forecasting. Our model effectively integrates both spatial dependency and temporal trend by enhancing the single time series to multiple dimensional based on a k-nearest neighbor graph. The embedding of turbine identity enables turbine-specific forecasts. The encoder-decoder structure overpasses the commonly used power curve transformation step in wind power forecasting problem, and improves the forecasting accuracy compared with classical approaches.

For future work, we would investigate the approach with probabilistic modeling to improve our model. This provides an uncertainty quantification about the forecasts, which will increase the interpretability of our model. The difficulty would be the probabilistic modeling of wind power. We plan to utilize the auto-regressive recurrent networks \cite{salinas2019deepar} and physical laws to tackle 
this problem.

\bibliographystyle{IEEEbib}
\bibliography{arxiv}

\end{document}